# Nano Version Control and "Robots of Robots" – Data Driven, Regenerative Production Code

Lukasz Machowski
Synthesis Software Technologies Johannesburg, South Africa
luke@synthesis.co.za
0000-0002-7011-659X

Tshilidzi Marwala
Department of Electrical and Electronic Engineering Science
University of Johannesburg
Johannesburg, South Africa
0000-0001-7372-5510

*Abstract*—A reflection of the Corona pandemic highlights the need for more sustainable production systems using automation. The goal is to retain automation of repetitive tasks while allowing complex parts to come together. We recognize the fragility and how hard it is to create traditional automation. We introduce a method which converts one really hard problem of producing sustainable production code into three simpler problems being data, patterns and working prototypes. We use developer seniority as a metric to measure whether the proposed method is easier. By using agent-based simulation and NanoVC repos for agent arbitration, we are able to create a simulated environment where patterns developed by people are used to transform working prototypes into templates that data can be fed through to create the robots that create the production code. Having two layers of robots allow early implementation choices to be replaced as we gather more feedback from the working system. Several benefits of this approach have been discovered, with the most notable being that the Robot of Robots encodes a legacy of the person that designed it in the form of the 3 ingredients (data, patterns and working prototypes). This method allows us to achieve our goal of reducing the fragility of the production code while removing the difficulty of getting there.

## I. INTRODUCTION

In the book: "Automation and collaborative robotics" [1], the authors make a compelling conclusion that "work in the future will change for many people and some areas of employment will be radically different over time". With the advent of the global Corona-19 Pandemic [2], this statement has just become a reality. In this paper, we explore how this has become true, and how Nano Version Control [3] can be used to effectively create an agent-based simulation system that enables us to create sustainable automation for the future. We demonstrate a practical way in which "process automation can start moving into process optimization", which according to Serge Mankovski [1], is the next meaningful step for automation and collaborative robotics. The final thought for the book [1] gives us a powerful vision where people do not need to compete with machines, but rather "should be more human". This paper gives us a concrete and practical example of how we can do this. The novelty in the approach is to use two layers of software robots and to split one hard problem of maintaining production code into three simpler problems, being data, patterns and working prototypes. To our knowledge, we are the first to propose the two-layer software robot approach and to utilize the data, patterns and prototype triad to simplify the creation of regenerative production code.

### A. Software Robots and Prior Research

The IEEE defines "a robot is an autonomous machine capable of sensing its environment, carrying out computations to make decisions, and performing actions in the real world." [4]. This broad definition can be applied to physical robots in the aerospace, consumer, disaster-response, drones, education, entertainment, exoskeletons, humanoids, industrial, medical, military & security, research, self-driving-cars, telepresence and underwater [5].

Software robots are not traditionally associated with this definition of robots, however the increased popularity of Robotic Process Automation (RPA) [1] begins to position the software robot as a meaningful analog to the traditional robots [5]. Interestingly, Wooldridge and Jennings, some of the early influencers of agent-based systems, described these as "softbots" (software robots) [6].

Agents are characterized by having autonomy, social ability, reactivity, and pro-activeness [6]. Sometimes, the additional characteristics of mobility, veracity, benevolence, rationality, and learning are also bestowed onto agents [6][7], making them closer to what we imagine physical robots to be.

In this paper, we chose to define a category of software robots that resemble agent-based systems but ones that interact with the DevOps world, typically encountered in the cloud. The software automation described in this paper has sufficient similarity to that of physical robots and software agents in the way that they sense and interact with each other through the software environment. For simplicity, in the rest of the paper, we will refer to "software robots" simply as "robots".

## II. WORKING FROM HOME AND THE CORONA-19 PANDEMIC

This section provides a background narrative to the inspiration for this approach. Although the method would still be valid whether the pandemic happened or not, it is important to understand the circumstances under which the approach became obvious because it is in these circumstances that we may come to understand the positioning and value of the method. Furthermore, it highlights how a drastic change in the way people work across the world was needed to necessitate the need for the method.

### A. Changes to the Ways of Working

Thinking back 2 years from the time of writing, in 2019, the daily routine for a typical software developer in South Africa would have resembled what is shown in Fig. 1. The context relates to producing sustainable software systems in a typical software development company.

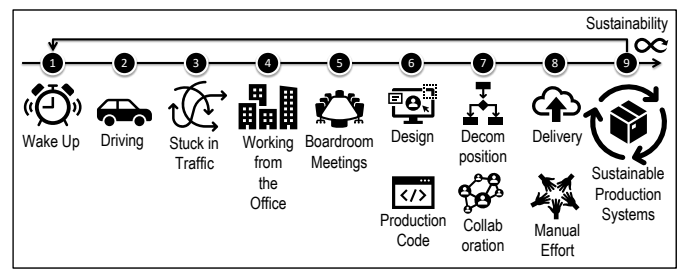

Fig. 1. T-2: Pre-Corona Work Cycle

This process was sustainable in the sense that it could be repeated indefinitely and would produce incremental improvements to the desired production system. Not all steps would have been relevant each day, but the flow captures the high-level themes that would be appropriate over long periods of time. For ease of reference, we will call this work cycle T-2 to indicate that it represents the work cycle from minus 2 years ago.

Then the Corona-19 Pandemic hit the world in 2020 [2]. This forced many industries to rethink the way in which people could do their work, and "Work from Home" became a reality. Software development was no exception. Things had to change as shown in Fig. 2 below. We refer to this period as T-1.

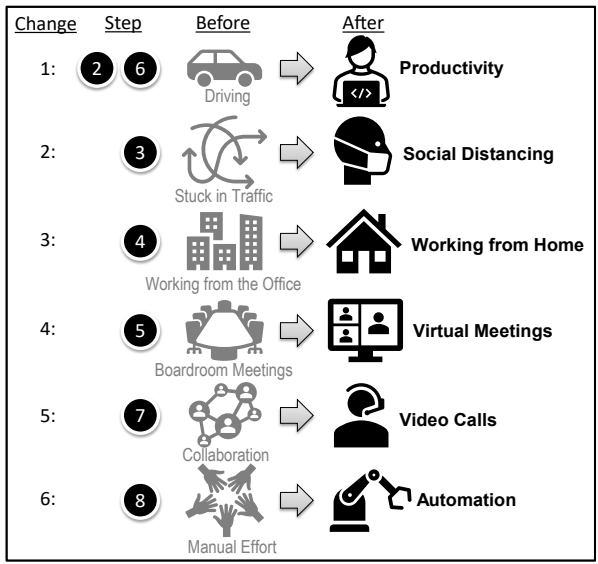

Fig. 2. Work from Home Changes

An explanation of each change is listed below along with the T-1 Corona Pandemic Work Cycle shown in Fig.3.

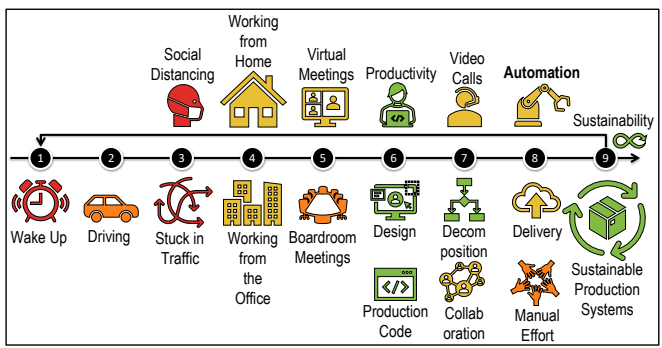

Fig. 3. T-1: Corona Pandemic Work Cycle and T0+ RAG Status

The **manual effort** started to be **automated** in step 8 of the work cycle. Since moving the work to the home created an opportunity for some workloads to be moved to the cloud, some people used this opportunity to start automating some of their workloads in that transition. Although the need for manual intervention was far from solved, the more important step was that more people started to realise the new opportunities for automation in their new environments working from home. **It is this opportunity for automation that was the trigger for making the proposed approach obvious.**

What we learnt from the T-1 work cycle was that it too was completely sustainable for software development. The rebalancing of various positives and negatives across the work cycle was generally reported as a net positive, despite there being new challenges that it introduced. This has an interesting effect when it comes to deciding what way of work we go back to once the Corona Pandemic is over.

*B. The Current Work Cycle*

This brings us to T0, right now (2021, at the time of writing). Although the Corona Pandemic is nowhere near being finished, several people are starting to think about how the world of work will look after the pandemic. There is no doubt that we will not be going back to the same way of working as T-2, however, we do have an opportunity to re-assess the ways of working and possibly keep the parts that we like and remove the parts that we do not like.

### III. WORK CYCLE SENTIMENT

A poll was conducted with software developers working on production systems using the T0 work cycle. Their sentiment is shown in Fig. 3, as coloured items with green being good, yellow being okay, orange being poor and red being bad, thus giving us a Red-Amber-Green (RAG) status.

We highlight some of the interesting sentiments that emerge for automation. There are two positive sentiments and two negative sentiments.

P1) We want to **automate repetitive tasks** because that is what software robots have been shown to do well [1] [8]. In fact, the recent popularity of Robotic Process Automation is an indicator of how much opportunity this approach can unlock [8].

P2) We want to see a **complex output come together** through **automatic composition of simpler parts**. The phrase "A million things needed to go perfectly for that to work" comes to mind. Software developers report a great sense of satisfaction [1] when a long and non-trivial automation completes at the touch of a button. We want to keep this.

N1) In this study, we have seen that creating software robots, and the associated **automation can be very fragile**. There is a critical mass that needs to be achieved before the automation "sticks". If the effort is stopped before the tipping point, then all the effort already invested is effectively lost and we tend to "go back to old habits". However, if the automation is pushed through and one overcomes the tipping point, the automation is likely to sustain for many years to come, paying itself back many times over. This non-linearity in automation effort-vs-reward is frustrating because it is hard to predict whether the automation will pay itself back. Our approach in this paper addresses part of this concern.

Part of this study elicited an interesting relationship around the critical mass curve of automation effort versus the perceived reward is shown in Fig. 4.

N2) We found that it is **extremely hard to get this automation right**, because of factors as mentioned in Fig. 4, but also because it is a highly creative process that requires ingenuity and pattern recognition to spot the opportunities for effective automation. This is not a trivial task. This paper describes how we retain the need for humans-in-the-loop to precisely do this task, because they are effective at it. As Artificial Intelligence techniques evolve, they will be able to augment people in this job [1][8][9].

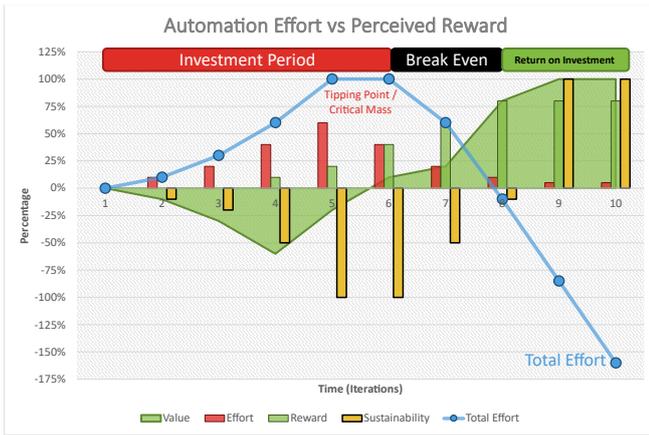

Fig. 4. Automation Effort versus Reward Relationship

This relationship should still be studied as further research.

The following is a description of the various sections of Fig. 4. We can see that a large amount of automation effort is required as input before any perceived reward is felt. In the early stages of automation (time 1-4), the automation initiative is highly unsustainable, meaning that if the automation were to be used in that manner (without improvement) for an extended period of time, it would result in a negative impact on productivity, thus making it unsustainable. There is a tipping point (at time 5 and 6) where the total effort into the system maxes out because the effort starts to decrease but the reward starts to increase. We have found that stopping the automation effort anywhere before this tipping point results in us slipping back to old habits and the overall initiative is seen as having destroyed value. There is a lag between the tipping point (time 6) and when it is sustainable (time 8) because people need the opportunity to build trust in the automation. Eventually (time 9), enough trust has been built that the automation is deemed by people to be sustainable, and this is usually once the sustained reward has more than paid back to total effort that was invested into the automation. Little literature could be found describing this relationship so this may be an area of further research to understand this non-linear dynamic more closely. The Robotic Process Automation Handbook [8] mentions this around the Return On Investment (ROI) but doesn't describe the dynamic mentioned above.

## IV. DESIGNING SUSTAINABLE SOFTWARE AUTOMATION

The previous section ended off with observations around the current state of automation, and the non-linear dynamic that needs to be overcome in order to make software automation sustainable. Our goal is to preserve, and magnify, the traits that are already good about automation, while removing the negative traits that hinder us from having sustainable automation. This is shown in Fig. 5.

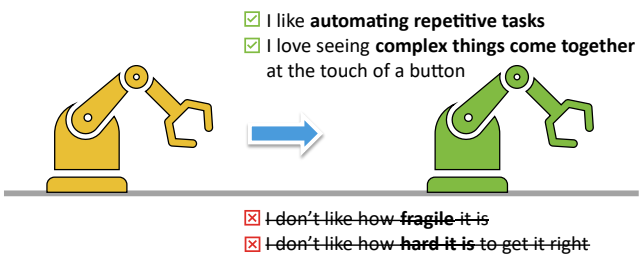

Fig. 5. Making Automation Sustainable

The two traits to preserve are:

Good Trait 1) **Automation of repetitive tasks** – Encoding the instructions to automatically reproduce repetitive tasks has been the hallmark of robotic automation for many years [11].

Good Trait 2) **Composition of complexity at the touch of a button** – seeing many pieces come together to form a complex, but useful output, is extremely rewarding and satisfying. As we build trust in the automation developed, more and more of the complexity can be built up using robots.

The two traits that must be reduced, or better yet, removed are:

Bad Trait 1) **Fragility** – The automation tends to be fragile. It's not enough just to make it robust, we also need to make it antifragile [10]. One way of dealing with fragility is to protect it from difficulties. This has proven unsustainable in software. The other way is to do the opposite and design the software system in a way that allows it to "benefit from shocks, grow when exposed to volatility, randomness, disorder and stressors" [10]. This paper presents a way that takes this idea to an extreme, where we ultimately consider the production source code as disposable, allowing us to create and destroy it cheaply and quickly so that we can let it progressively get stronger by disposing of previous iterations that made it weak. There is growing support for this antifragile thinking with a proposal for an antifragile software manifesto that is disrupting the agile development space [11]. Disposable production code is one way to think of the outcome that we are aiming for, but probably a better term for it might be "Regenerative Production Code". This embodies the idea that the production code can regenerate as needed, which implies varying degrees of tolerance to it being disposed of and recreated. To the authors knowledge, this would be a new term that we have not yet heard of to describe this type of approach to production software code.

Bad Trait 2) **Hard to get it right** – It is very difficult to overcome the non-linear dynamic described in Fig. 4, so it is important that we find a way to convert one really hard problem into simpler ones. This paper presents a practical way in which we can convert one hard problem into three simpler ones. We will show later that these 3 simpler problems are "**data**", "**patterns**" and "**working prototypes**".

### A. A System Diagram for Robots of Robots

The system diagram shown in Fig. 6 shows the way that we can move from the old way of doing automation, which tends not to be sustainable, into a new way, which is sustainable.

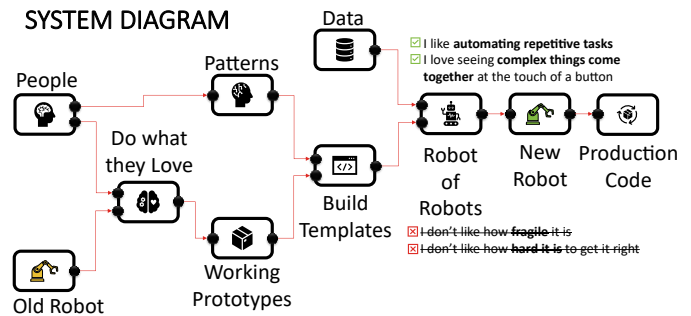

Fig. 6. Robots of Robots System Diagram

The method goes as follows: **People** must **do what they love doing** (programming, breaking complex problems down into simpler ones, composing them up and finding solutions to hard problems). They can take inspiration from the **old robot** (automation) and create a **working prototype** which captures the essence of the solution. This working prototype focusses on the mechanics of the solution (implementation) but uses valid labels for sections of the code that will substituted and multiplied out by the data. It's very important that at this stage we recognize that the labels are still valid in the language that the code is being written but is easy enough for patterns (usually regular expressions) to identify and replace with the data. The person then spots the **patterns** in the intended solution and authors the code that can **build the templates** from the working prototypes. This step is critical to the method because it departs from traditional meta-programming where the person needs to author the templates by hand, usually in a higher-level templating language. This traditional process is tedious and hard because the code they are authoring is not valid code in the language that it will be rendered to, and therefore requires specialized tooling to support the person when developing those templates. Instead, with the new method, the person is writing the patterns that transform the working prototype into a valid template in whatever templating language is chosen. We have found that using a combination of templating languages (multi-layered parameter substitution) is an effective technique because simple parameter replacement can be done using simple template syntax (eg: *{{variable}}* in Mustache templates) and complicated replacements can be implemented using more powerful syntax (eg: *${order.items[0].getTotal()}* in Apache Velocity). The **Robot of Robots** then takes the **data** for the problem domain and merges it through the generated templates to produce the new robot. This robot is then able to produce the **production code** that will deliver the solution to the problem. This approach means that both the new robot and final production code is disposable, and it instead puts the importance on the data, patterns and working prototype. This is what we mean by regenerative production code: the production code can be regenerated from data, patterns and working prototypes.

### B. Converting 1 hard problem into 3 simpler problems

Creating, maintaining, and regenerating **production software code** is hard to build, hard to change, painful to lose and expensive (both in time and in money). This is a hard problem to deal with. The proposed Robots of Robots method shifts the focus onto **data**, which is easy to create, easy to change and is getting easier to extract thanks to modern data tools and upcoming AI techniques. **Working prototypes** are easy to make, because people love building them, easy to run, easy to test and most importantly, easy to change because they are small and cheap to recreate with new technology stacks. The **patterns** are easy to test once you have them but difficult to write, which is exactly where keeping the "human-in-the-loop" [1] is important. We can use people to do what they are great at, which is spotting patterns. This means that we can effectively replace the one hard problem (building sustainable production code) with three simpler problems (data, patterns and working prototypes). We have found that this gives us a net-reduction in the effort required for that production code and a gain in the net-maintainability of the code, mainly because it is disposable, or more accurately, because it can be regenerated.

### C. Why two layers of robots?

Something characteristic of this design is that we have two layers of robots. Firstly, the robot that produces the desired output. Secondly, the Robot of Robots, which builds the first robot. This may seem counterintuitive at first.

To explain, we need to imagine a traditional mass-production factory, which may have a production-line surrounded by automated robots. Imagine a modern car being assembled in this factory. You might imagine that once you have the production line of robots, it would be unnecessary to add another layer of robots behind them, but it's not. The problem with only having one layer of robots is that it's extremely expensive and time consuming to create a factory, and then later, change the factory.

Typically, the process engineers would need to decide on key design metrics upfront when they know the least about the nuances of the end-to-end process. This can be mitigated by extending the analysis phase or reducing the unknowns in the solution, however with the pace at which widgets need to get to market, this is an opposing force to the time-to-market. Since the timescale is usually long for creating an automated production line like this, the choice of robots is done upfront, which means that by the time the production line produces its first output, there is a chance that some of the robots might already be outdated.

If we imagine instead that we are using software robots to build the production code, the pace at which the software industry changes, it's almost guaranteed that at least some of the robots in the production line will be out of date by the time the entire complex automation pipeline is operational.

Imagine instead, that you invested in a cheaper way to automate the creation of the pipeline of robots that do the work. This means that you can cheaply substitute and dispose of robots that were introduced early in the process with more effective robots once you know more about the actual end-to-end process. By making the robots disposable, you can regenerate parts of the production line with improved replacements, or alternatively, with more instances of the robots to balance the total throughput of the system.

### D. Cascaded code generation

When implementing complex production code bases, which is typical of low-level messaging protocols in the FinTech space, we have found that a very useful technique is to have multiple passes of the code generation. The output of the first code generation is required to write the working prototypes for the second layer of code generation, and so on. This creates a cascading design where one phase is the input for the subsequent phases. As an example, for messaging protocols, the first layer is used to create the specific messages for the protocol. The working prototype is a template for just one message with one field for the data. Each message along with every field for the message is captured as the data. The patterns are created by people to convert the working prototype of a message into a templating language where the data can be fed through to create all the messages that are required for the protocol. This is the first layer of the cascading code generation. The next layer is to create the activities that define the dynamic aspects of the messaging protocol (the sequence diagrams of the protocol). Having generated messages from layer 1 means that working prototypes of the activity code can be hand authored with standard tools because they reference the full library of available messages. An important part of this architecture is where partial implementations can be authored for the working prototypes, and these are merged (fused) into the production code by mixing it into the generated output using the template. This cascading design is repeated for a third phase for creating the domain models (entities) and finally the fourth phase for the synchronization code that determines the appropriate phase 2 activities to activate depending on the changes in the phase 3 model state.

## V. IMPLEMENTING ROBOTS OF ROBOTS

This section details the implementation approach for the system diagram shown in Fig. 6. The problem is well suited to implementation as an agent-based simulation [12] where additional challenges of arbitration between agents are solved using Nano Version Control repositories [3].

### A. Agent-Based Simulation

Agent-based simulation has multiple software agents interacting with each other through a simulated environment [12]. Typical examples of this approach are the swarming behavior of ants which can coordinate the finding and retrieval of a food source back to their nest by leaving behind a pheromone trail for other ants to follow. This is a great example where there is no centralized command-and-control structure governing the swarm, but rather simple rules used by each agent. The system-wide emergent behavior seems more complex than the sum of its parts.

A typical problem that arises in agent-based simulation is when one agent wants to modify the environment one way and another agent wants to modify it in an opposing way. This leads to a need for arbitration so that the simulation can choose which agent wins. This is undesirable because it introduces an unwanted bias that pre-supposes a solution.

### B. Nano Version Control Repos for Arbitrartion

The Nano Version Control repository [3] has been used to eliminate this arbitration bias in agent-based simulation by creating a new branch in the simulation repo every time agents' conflict. The approach is to have a branch created for each agent from the previous simulation state and make the agent think that it's the first to go. Each agent branch is then merged with every other agent and if any merge conflicts arise, a new simulation branch is created for every permutation of agents that conflicted, as shown in Fig 7. This leads to a design that naturally produces multiple solutions for one simulation. At first we were concerned that this strategy would generate thousands of solutions, but we found for the problem conducted in this study, that most agents agree, and merge conflicts only occur when there are genuinely multiple solutions to the problem.

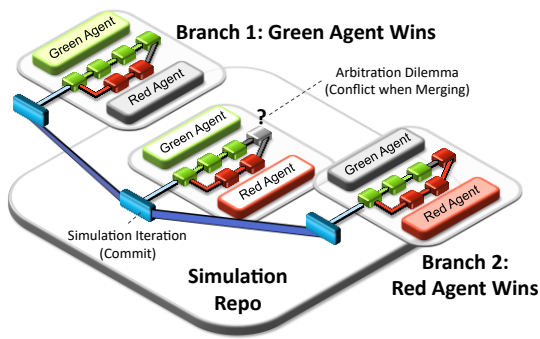

Fig. 7. New branches of simulation for arbitration of conflicts.

The system diagram shown in Fig. 6 is implemented as agents using the approach described in Fig. 7. This gives us the added benefit that a simulation of the production code can be previewed before doing the full run. In addition, if there is any point in the process where the Robots of Robots implementation generates conflicts in the simulated environment, we get multiple solutions for each arbitration solution, which is an interesting and powerful by-product of the approach.

## VI. IMPLICATIONS OF ROBOTS OF ROBOTS

The implication of this design is that we have converted one really hard problem into 3 simpler problems, as shown in Fig. 8.

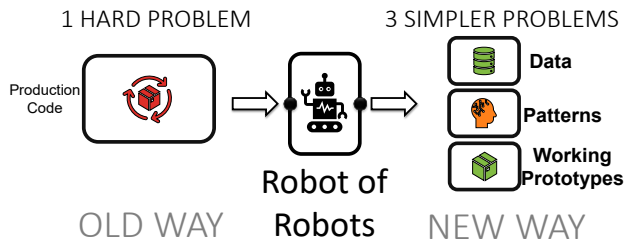

Fig. 8. Implication of Robots of Robots.

Since the data can be transported over long distances and times easily, with the option of sending partial data for proof-of-concept runs, the patterns can be encoded efficiently and transported with unit tests that prove that they work, and the working prototypes are small, this means that large production code bases can be transmitted very efficiently over long distances and times. With the advent of space missions to distant planets like Mars, it becomes more important to understand data structures that can be tolerant to such space-time practicalities. This implies that people working on the production system can be in different places at different times and their work can effectively be merged using standard Version Control principles [3].

This approach to regenerative production code extends the shelf-life of a system because major changes to the technology stack can be introduced and de-risked with overhauling the working prototype and appropriate patterns but keeping the data the same.

Since all 3 ingredients to the approach (data, patterns and working prototypes) can have associated unit tests, automated application of updates to the working prototype can be applied in the simulation before executing and unit tests can be used to verify the appropriateness of the update to the whole system. If the simulation and unit tests pass, then an automated update can be proposed by the system for a human to judge whether they agree to apply the changes. This leads to systems that can maintain themselves for a well-defined set of boundary conditions. The benefit to people is that for obvious cases, the machines can keep themselves running and people only need to provide feedback occasionally if the context moves out of that set of boundary conditions.

Since all 3 ingredients can also be version controlled, it means that the production system can be rolled back to previous states if something goes wrong. This is useful in scenarios where a person that was unable to monitor the auto-updating process described above but they later find that they prefer to roll back to a previous state for strategic concerns.

An extension of the NanoVC way of thinking [3] is that you can also consider "alternate futures" where a proposed change to the system is implemented in a branch and the impact to the overall system can then be simulated before doing the change for real. This mitigates a lot of risk associated with updating production code bases.

The Robot of Robots is in essence, an encoding of the designer of the production line of robots. This means that the **Robot of Robots is a legacy of the person that designed it**. This gives us a new perspective on how we can encode artificial intelligence in terms of 3 main ingredients (data, patterns and working prototypes). This eliminates the key-man dependency that is typically associated with large production systems, allowing that person to move on to other projects or companies. The unit tests associated with the data, patterns and working prototypes are an explicit feedback loop which regression proof the original intention of that person.

## VII. Research Methodology

The method for conducting the research was to compare the seniority of developers who could sustainably maintain and change a production software system, before and after using the proposed method.

The system consisted of a full stack application for a messaging protocol in the FinTech space. Cascaded code-generation was needed to implement the messages in phase 1 and the associated activities for the protocol in phase 2. A non-trivial set of changes to the messages and activities was the benchmark to measure what seniority of developer could make the required changes to the system. A fixed timeframe was given for the changes to be made, both using the old method and using the new method. This is the basis for measuring "harder" or "easier" for the purposes of this paper. If a lower seniority of developer is capable of making the required changes then we deem the difficulty to be "easier".

To simplify the seniority levels, we are simply defining the levels to be "**junior**", "**intermediate**" and "**senior**" developers. A **junior** developer would be someone that is still unconsciously incompetent in their job and frequently requires input from someone more senior to assist them and give them direction. An **intermediate** developer is consciously competent of the job they are required to do but must still be given very clear instructions of the tasks to complete. The **senior** developer is consciously competent and has enough managerial experience to be able to provide the required direction and guidance for the intermediates and juniors. They also have enough technical foresight to drive the necessary changes and make difficult judgement calls that are needed for production code bases.

The following numbers of developers were included in this study, as shown in Table 1:

TABLE I. Seniorities and Number of Developers

| Juniors (Jun.) | Intermediates (Int.) | Seniors (Sen.) |
|---|---|---|
| 2 | 3 | 1 |

## VIII. Results

### A. Measurement

Using the traditional software development lifecycle to change and maintain the production source code, we see that only the senior developer was able to make the necessary changes on their own, as shown in Table 2.

TABLE II. Seniority of Developers Capable of Making Changes using the Old method

| Task | Jun. | Int. | Sen. |
|---|---|---|---|
| Change Production Source Code | 0 | 0 | 1 |

When we used the proposed method to break the problem down into the three sub-problems, we get the results shown in Table 3.

TABLE III. Seniority of Developers Capable of Making Changes using the New method

| Task | Jun. | Int. | Sen. |
|---|---|---|---|
| Encode Required Changes as Data Changes | 2 | 3 | 1 |
| Encode Required Pattern Changes | 2 | 3 | 1 |
| Update Working Prototype | 2 | 3 | 1 |

### B. Analysis

The results show that once the new approach was explained to the developers, all 6 developers at all seniorities were able to perform the required changes, as opposed to only the senior developer being able to perform the required changes in the given timeframe using the old method.

This small sample shows that it is indeed much simpler to use the new method to make the required production source code changes because now junior developers can be used where previously only senior developers where capable of making the changes.

When we say that we convert 1 hard problem into 3 simpler ones, we understand that data, patterns and prototypes in general may have their own massive complexities, but for the purposes of how we apply these 3 sub-problems in this approach, we are only relying on the basic aspects of all 3 of those sub-problems. Hence, we confidently say that they are 3 simpler problems, for this problem-space.

We recognize that these are only early measurements on a very small sample, so further work will need to be done on more diverse production systems, with many more developers. However, the startling result in the reduced developer seniority shows that this method can have a significant financial and practical impact on the way certain production code bases can be maintained. These promising results warrant further investigation into this method.

## IX. Conclusion

We have introduced a method which converts one really hard problem of producing sustainable production code into three simpler problems being data, patterns and working prototypes. By using agent-based simulation and NanoVC repos for agent arbitration, we are able to create a simulated environment where patterns developed by people are used to transform working prototypes into templates that data can be fed through to create the robots that create the production code. Having two layers of robots allow early implementation choices to be replaced as we gather more feedback from the working system. Several benefits of this approach have been discovered, with the most notable being that the Robot of Robots encodes a legacy of the person that designed it in the form of the 3 ingredients (data, patterns and working prototypes). This method allows us to achieve our goal of reducing the fragility of the production code while removing the difficulty of getting there.